\documentclass[sigconf]{acmart}

\AtBeginDocument{%
  }

\usepackage{algorithmic}
\usepackage{algorithm}

\setcopyright{acmlicensed}
\copyrightyear{2026}
\acmYear{2026}
\acmDOI{XXXXXXX.XXXXXXX}
\acmConference[ACM MM '26]{Proceedings of the 30th ACM International Conference on Multimedia}{October 2026}{Seattle, WA}
\acmISBN{978-1-4503-XXXX-X/2026/10}

\begin{document}

\title{AMR: Adaptive Modality Routing for Multimodal Polyglot Speaker Identification}

\author{Chuxiao Zuo\textsuperscript{*}}
\email{zuochuxiao@honor.com}
\affiliation{%
  \institution{Honor Device Co., Ltd}
  \city{Nanjing}
  \country{China}
}

\author{Yao Zhu\textsuperscript{*}}
\email{zhuyao2@honor.com}
\affiliation{%
  \institution{Honor Device Co., Ltd}
  \city{Shanghai}
  \country{China}
}

\author{Minqiang Xu\textsuperscript{*}}
\email{xuminqiang@honor.com}
\affiliation{%
  \institution{Honor Device Co., Ltd}
  \city{Shanghai}
  \country{China}
}

\author{Manhong Wang}
\email{wangmanhong@honor.com}
\affiliation{%
  \institution{Honor Device Co., Ltd}
  \city{Shanghai}
  \country{China}
}

\author{Yunke Zhang}
\email{zhangyunke@honor.com}
\affiliation{%
  \institution{Honor Device Co., Ltd}
  \city{Nanjing}
  \country{China}
}

\author{Fei Huang\textsuperscript{$\dagger$}}
\email{huangbo6@honor.com}
\affiliation{%
  \institution{Honor Device Co., Ltd}
  \city{Shanghai}
  \country{China}
}

\renewcommand{\shortauthors}{Chuxiao Zuo et al.}

\begin{abstract}
  Multimodal speaker identification systems face two key challenges in real-world deployment: missing modalities and language mismatch between training and testing conditions.  In practical scenarios, background multi-speaker conversations, ambient noise, and overlapping speech further degrade identification accuracy.  To address these challenges, we propose a multimodal polyglot speaker identification system for the POLY-SIM 2026 Grand Challenge. The system is fundamentally built upon \textbf{Adaptive Modality Routing~(AMR)}, a modality fusion module that dynamically assesses per-sample input quality and integrates modality information. Specifically, AMR employs two modality adapters to process the embeddings extracted from a linguistically robust audio encoder~(W2V-BERT 2.0) and a large-scale pretrained face encoder~(IResNet-18), producing modality-adapted embeddings. Based on these adapted embeddings, a trainable router estimates dynamic modality weights, which are subsequently applied to aggregate the modality-specific logits for the final prediction. To optimize this routing mechanism, we adopt a modality-aware training strategy that constructs four types of sample pairs to simulate diverse input conditions, with KL divergence serving as explicit supervision for weight assignment.  Experimental results on the POLY-SIM 2026 evaluation set show that the proposed system achieves identification accuracy of 99.93\%~(English multimodal, P3), 100.00\%~(Urdu multimodal, P5), 97.50\%~(English audio-only, P4), and 98.83\%~(Urdu audio-only, P6).  The average accuracy across all four protocols is 99.07\%, surpassing the Fusion and Orthogonal Projection~(FOP) baseline by 32.73\%.
\end{abstract}




\maketitle
\let\thefootnote\relax
\footnotetext{\textsuperscript{*}These authors contributed equally to this work.}
\footnotetext{\textsuperscript{$\dagger$}Corresponding author.}

\section{Introduction}
While audio-only speaker identification~\cite{ivector, xvector, ecapa, spkin} often degrades under real-world noise and linguistic variations, integrating voice and face modalities offers a robust alternative. This multimodal synergy is crucial for polyglot speaker identification, as facial identity provides a stable, language-invariant anchor that complements fine-grained vocal characteristics~\cite{fv5}. Driven by advancements in deep representation learning~\cite{wav2vec2, w2vbert, arcface, webface4m} and the FAME challenge series~\cite{mavceleb, fame2024, fame2026}, recent systems have achieved remarkable multilingual face-voice association through techniques like data quality filtering~\cite{hlt2024}, cross-contrastive learning~\cite{fameav2024}, and Fusion and Orthogonal Projection~(FOP)~\cite{fop}. However, these prior efforts strictly assume the availability of both modalities during inference. In practical deployments, modalities are frequently missing due to visual occlusion, sensor failure, or privacy constraints---a critical vulnerability that severely degrades conventional multimodal performance~\cite{missingmodality}.

The POLY-SIM 2026 Grand Challenge~\cite{polysim2026eval} introduces missing modality by withholding face images at test time in the closed-set identification setting. The challenge defines four evaluation protocols: in-language multimodal~(P3), in-language audio-only~(P4), cross-lingual multimodal~(P5), and cross-lingual audio-only~(P6). Beyond these challenge-defined settings, real-world deployment introduces additional interference that further degrades identification accuracy, including overlapping speech from background multi-speaker conversations, audio degradation caused by ambient noise, and reduced face image quality due to camera motion or poor lighting.  The FOP baseline~\cite{fop}, representative of current multimodal methods, achieves 97.44\% on English multimodal~(P3) and 98.48\% on Urdu multimodal~(P5), yet only 37.75\% on English audio-only~(P4) and 31.70\% on Urdu audio-only~(P6). This demonstrates that existing methods, designed under the assumption of complete modalities, fundamentally cannot handle the combined challenges of missing modality and unreliable input quality.

To address these challenges, we propose a multimodal polyglot speaker identification system for the POLY-SIM 2026 Grand Challenge.  Our main contributions are as follows:
\begin{sloppypar}

\begin{enumerate}
  \item We propose Adaptive Modality Routing~(AMR), a modality fusion module that dynamically assesses per-sample input quality and integrates modality information. Specifically, AMR employs two modality adapters to process the embeddings extracted from a linguistically robust audio encoder and a large-scale pretrained face encoder, producing modality-adapted embeddings. Based on these adapted embeddings, a trainable router estimates dynamic modality weights,  which are subsequently applied to aggregate modality-specific logits for the final prediction. To optimize this routing mechanism, we adopt a modality-aware training strategy that constructs four types of sample pairs to simulate diverse input conditions, with KL divergence serving as explicit supervision for weight assignment.

  \item We develop two independently optimized modality encoders, supported by a rigorous data preparation pipeline. Specifically, the audio encoder is fine-tuned via a three-stage progressive strategy equipped with Multi-Frame Aggregation~(MFA), while the face encoder leverages robust pretraining on WebFace4M~\cite{webface4m}.
  \item Experimental results on the POLY-SIM 2026 evaluation set show that the proposed system achieves identification accuracy of 99.93\% (English multimodal, P3), 100.00\% (Urdu multimodal, P5), 97.50\% (English audio-only, P4), and 98.83\% (Urdu audio-only, P6). The average accuracy across all protocols is 99.07\%, surpassing the FOP baseline~\cite{fop} by 32.73\%.
\end{enumerate}
\end{sloppypar}

\section{Related Work}
\label{sec:related}

\textbf{Speaker Recognition.}
Deep speaker recognition systems have evolved from i-vector~\cite{ivector} and x-vector~\cite{xvector} frameworks to end-to-end neural architectures, including ECAPA-TDNN~\cite{ecapa} and ResNet-based models~\cite{resnet}.  Recently, self-supervised learning~(SSL) models such as Wav2Vec 2.0~\cite{wav2vec2} and W2V-BERT 2.0~\cite{w2vbert} pretrained on large-scale unlabeled speech have demonstrated strong speaker-discrimination capability when fine-tuned with adapter modules~\cite{adapter}.  Our audio encoder builds on W2V-BERT 2.0 with a custom MFA architecture that aggregates information across all transformer layers, capturing multi-scale acoustic information beyond what the final layer alone can capture.

\textbf{Face Recognition.}
Recent advances in face recognition are driven by both loss function design and backbone architecture.  On the loss side, ArcFace~\cite{arcface} introduced additive angular margin loss, MagFace~\cite{magface} leverages feature magnitude as a quality indicator, and AdaFace~\cite{adaface} adapts the margin based on image quality.  On the backbone side, the IResNet family~\cite{arcface}---including IResNet-18, IResNet-50, and IResNet-100---has been the most widely adopted CNN architecture, commonly trained on large-scale datasets such as WebFace4M~\cite{webface4m}.  More recently, Vision Transformer~(ViT) backbones have been introduced for face recognition, with TransFace~\cite{transface} demonstrating that pure Transformer architectures can match or exceed CNN performance on standard benchmarks through dynamic patch sampling and attention enhancement.  ConvNeXt-based~\cite{convnext} face recognition models have also emerged as modern CNN alternatives, bridging the design gap between CNNs and Transformers.  Hybrid CNN-Transformer architectures that combine local feature extraction with global context modeling represent another promising direction.  Despite these advances, the IResNet family remains a robust and computationally efficient choice, particularly for tasks with a limited number of classes.  In our system, we adopt IResNet-18 pretrained with AdaFace on WebFace4M.  Our experiments showed that the larger IResNet-100 backbone performed worse than IResNet-18 on the 70-speaker task, so we opted for the lighter model.

\textbf{Face-Voice Association.}
Establishing the association between faces and voices has attracted growing research interest since the introduction of the VoxCeleb1 dataset~\cite{voxceleb}. Early work explored cross-modal mapping via deep latent space learning~\cite{fv4}. Subsequent methods developed structured fusion strategies, including single-branch multimodal training~\cite{fv3}, precise alignment with enhanced gated feature fusion~\cite{fv2}, attention-based fusion with FOP~\cite{fop}, and face-voice association with inductive bias for maximum class separation~\cite{fv1}. The FAME challenge series~\cite{fame2024, fame2026} extended face-voice association to multilingual settings, introducing the MAV-Celeb dataset~\cite{mavceleb} with language annotations. However, these methods assume both modalities are available during inference.

\begin{figure*}[t]
  \centering
 \includegraphics[width=0.95\linewidth]{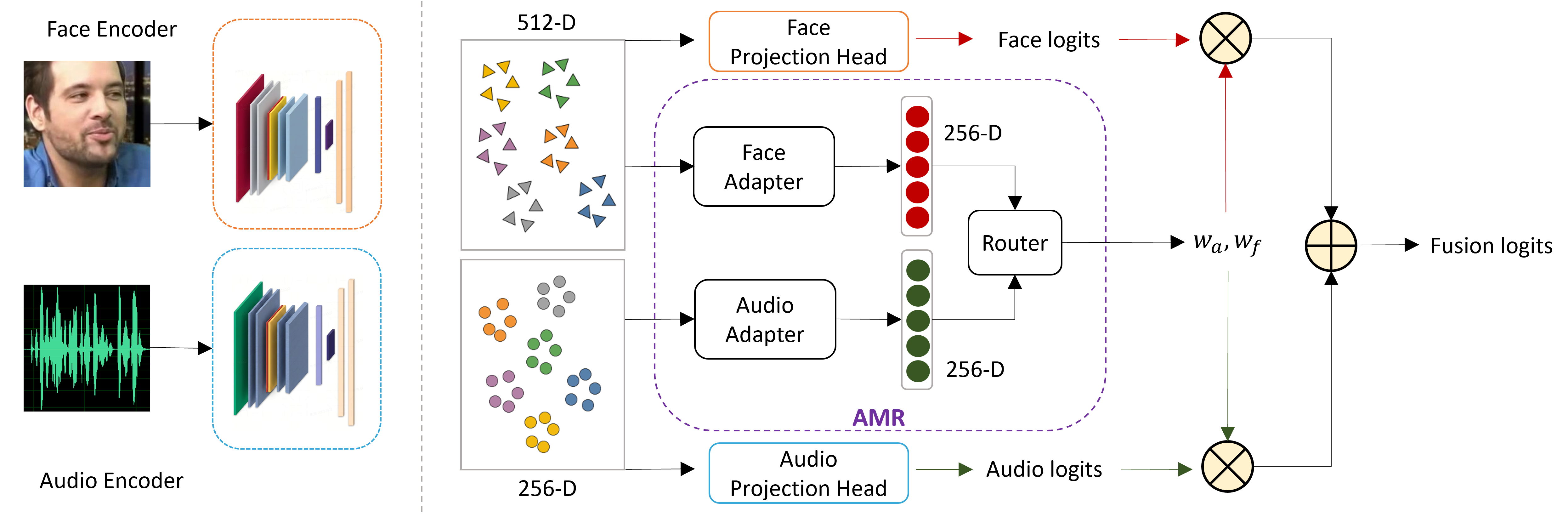}             
  \caption{Overview of the proposed multimodal polyglot speaker identification system with AMR module.}
  \label{fig:architecture}
  \Description{Architecture diagram showing audio and face pathways converging at an AMR router module that produces modality weights for logit fusion.}
\end{figure*}

\section{Proposed Method}                
\label{sec:method}

Figure~\ref{fig:architecture} illustrates the proposed multimodal polyglot speaker identification system. The system comprises three components: an audio encoder~(Section~\ref{sec:audio}), a face encoder~(Section~\ref{sec:face}), and an AMR module~(Section~\ref{sec:amr}). The audio encoder takes raw waveform as input and produces a speaker embedding $\mathbf{e}_a$ and audio logits $\mathbf{l}_a$; the face encoder takes face imagery as input and produces a face embedding $\mathbf{e}_f$ and face logits $\mathbf{l}_f$. The AMR module first projects both embeddings into a common representation space via two modality adapters, and then employs a trainable router to estimate per-sample modality weights from the adapted representation. The final classification logits are obtained as a weighted combination of the modality-specific logits. Training is conducted in two stages: first, the encoders are trained independently on modality-specific data; then, the encoders are frozen while only the AMR module is trained. It is worth noting that a rigorous data preparation pipeline~(Section~\ref{sec:data}) is used to construct high-quality training data from both the training set and open-source resources, which is crucial for the performance of both encoders. During the inference stage, for conditions without face imagery~(P4, P6), the face pathway is deactivated, leaving the system to rely solely on audio input.

\subsection{Audio Encoder}
\label{sec:audio}
\subsubsection{Architecture.}
The audio encoder takes a raw waveform $\mathbf{x} \in \mathbb{R}^{T}$ as input, where $T$ denotes the length of $\mathbf{x}$, and produces a 256-dimensional speaker embedding $\mathbf{e}_a \in \mathbb{R}^{256}$ as output. The encoder uses W2V-BERT 2.0~\cite{w2vbert} as a self-supervised frontend, which generates 25 layers of hidden states: one from the input embedding layer and one from each of the 24 transformer layers. We denote these hidden states as $\mathbf{H}_i \in \mathbb{R}^{L \times 1024}$ for $i = 1, \ldots, 25$, where $L$ is the number of frames, and 1024 is the hidden dimension.

To capture multi-scale acoustic information, the MFA architecture processes all 25 hidden states:
\begin{equation}
  \mathbf{e}_a = f_{\text{MFA}}(\mathbf{H}_1, \mathbf{H}_2, \ldots, \mathbf{H}_{25};\, \theta_{\text{MFA}}) \in \mathbb{R}^{256}
  \label{eq:mfa}
\end{equation}
Specifically, each hidden state passes through a modality adapter to reduce dimensionality, producing compact frame-level features. These adapter outputs are concatenated along the feature dimension and subsequently aggregated over time by Attentive Statistics Pooling into a fixed-length vector. Finally, a bottleneck layer projects this representation to the final 256-dimensional speaker embedding $\mathbf{e}_a$.

During training, an ArcFace~\cite{arcface} projection head maps the embedding $\mathbf{e}_a$ to audio logits $\mathbf{l}_a$.  Let $\mathbf{W} \in \mathbb{R}^{256 \times K}$ denote the weight matrix, where $K$ is the number of speakers.  The audio logit $l_{a,j}$ for class $j$ is:
\begin{equation}
 l_{a,j} = s \cdot \cos(\theta_j), \quad \theta_j = \arccos\!\Bigl(\frac{\mathbf{W}_j^\top \, \hat{\mathbf{e}}_a}{\|\mathbf{W}_j\|}\Bigr)
  \label{eq:arcface}
\end{equation}
where $\hat{\mathbf{e}}_a = \mathbf{e}_a / \|\mathbf{e}_a\|$ is the L2-normalized embedding, $\mathbf{W}_j$ is the $j$-th column of $\mathbf{W}$, and $s$ is the scale factor.  In vector form:
\begin{equation}
  \mathbf{l}_a = s \cdot \hat{\mathbf{W}}^\top \hat{\mathbf{e}}_a \in \mathbb{R}^{K}
  \label{eq:audio_logits}
\end{equation}
where $\hat{\mathbf{W}}$ denotes column-wise L2 normalization of $\mathbf{W}$.  An additive angular margin $m$ is applied only to the target class $y$: $l_{a,y} = s \cdot \cos(\theta_y + m)$.  The training loss is the cross-entropy computed over all class logits.

\subsubsection{Training Strategy.}
\label{sec:audio_training}

We employ a three-stage progressive training strategy.  Table~\ref{tab:audio_stages} summarizes the configuration of each stage.

\begin{table}[t]
  \caption{Three-stage progressive audio encoder training configuration.}
  \label{tab:audio_stages}
  \begin{tabular}{lccc}
    \toprule
    Stage & Trainable & LR & Epochs \\
    \midrule
    Stage 1 & Projection & $10^{-3}\!\to\!10^{-4}$ & 15 \\
    Stage 2 & MFA + Projection & $10^{-4}\!\to\!10^{-5}$ & 15 \\
    Stage 3 & MFA + Projection & $10^{-4}\!\to\!10^{-5}$ & 5 \\
    \bottomrule
  \end{tabular}
\end{table}

\textbf{Stage 1: Projection-only training.}
The model is initialized from a checkpoint pretrained on VoxCeleb2~\cite{voxceleb2} and VoxBlink2~\cite{voxblink2}, which includes pretrained W2V-BERT 2.0 and MFA weights.  Both the frontend and MFA remain frozen; only the ArcFace projection layer is trainable.  We use the AdamW optimizer with learning rate $10^{-3}$ and a cosine annealing schedule with warmup.  The model is trained for 15 epochs using fixed 5-second audio chunks.

\textbf{Stage 2: MFA fine-tuning.}
Initialized from the Stage~1 checkpoint, the model is trained for 15 epochs with the frontend network kept frozen, optimizing only the MFA layers and the ArcFace projection head. The learning rate is adjusted to $10^{-4}$ with a weight decay of 0.01. To enhance the model's robustness to varying utterance durations, variable-length inputs~(1--15 seconds) are used instead of fixed-length chunks.

\textbf{Stage 3: TTS augmented fine-tuning.}
The MFA and projection head are further fine-tuned using data augmented with Text-to-Speech (TTS) synthesis generated by CosyVoice3~\cite{cosyvoice3} and VoxCPM2~\cite{voxcpm}. These TTS models generate additional speech by cloning speaker characteristics from reference audio, with all synthesized data restricted to English in compliance with the challenge rules. A speaker similarity filter is further applied to retain only synthetic samples that closely match the target speaker. The model is initialized from the Stage~2 checkpoint and trained for 5 epochs with a learning rate of $10^{-4}$.

\subsection{Face Encoder}
\label{sec:face}

\subsubsection{Architecture.}
The face encoder takes a face image as input and produces a 512-dimensional face embedding $\mathbf{e}_f \in \mathbb{R}^{512}$.  The encoder uses {IResNet-18\textsuperscript{1}}~\cite{arcface} with Squeeze-and-Excitation~(SE) attention modules~\cite{senet}, initialized from a pretrained checkpoint on WebFace4M~\cite{webface4m} trained with the AdaFace~\cite{adaface} loss.

\footnotetext{\textsuperscript{1}https://github.com/deepinsight/insightface}

\subsubsection{Training Strategy.}
We fine-tune the face encoder on the 70 target speakers of the POLY-SIM 2026 challenge using a single-stage fine-tuning protocol.  The pretrained IResNet-18 is fully unfrozen and fine-tuned with the ArcFace~\cite{arcface} loss.  A projection head $\mathbf{W}_{\text{fc}} \in \mathbb{R}^{512 \times 70}$ maps the L2-normalized embedding $\hat{\mathbf{e}}_f = \mathbf{e}_f / \|\mathbf{e}_f\|$ to the face logits:
\begin{equation}
  \mathbf{l}_f = s \cdot \hat{\mathbf{W}}_{\text{fc}}^\top \hat{\mathbf{e}}_f \in \mathbb{R}^{70}
  \label{eq:face_logits}
\end{equation}
where $\hat{\mathbf{W}}_{\text{fc}}$ denotes column-wise L2 normalization of $\mathbf{W}_{\text{fc}}$, $s = 32.0$ is the scale factor, and an additive angular margin $m = 0.2$ is applied to the target class.  To address the imbalanced sample distribution across speakers, the dataloader employs \texttt{WeightedRandomSampler} to ensure balanced sampling during training.  Table~\ref{tab:face_training} summarizes the training configuration.

\begin{table}[t]
  \caption{Face encoder fine-tuning configuration.}
  \label{tab:face_training}
  \begin{tabular}{ll}
    \toprule
    Parameter & Value \\
    \midrule
    Loss function & ArcFace~(scale $s=32.0$, margin $m=0.2$) \\
    Optimizer & AdamW \\
    Learning rate & $5 \times 10^{-4}$ with cosine decay \\
    Batch size & 128 \\
    Epochs & 30 \\
    Input size & $112 \times 112$ \\
    Data augmentation & Random horizontal flip ($p=0.5$), color jitter \\
    \bottomrule
  \end{tabular}
\end{table}


\subsection{Adaptive Modality Routing}
\label{sec:amr}

A key challenge in multimodal fusion is that input quality varies across samples.  In practical scenarios, background multi-speaker conversations, ambient noise, and overlapping speech degrade audio quality; face imagery may depict a different identity or be entirely unavailable; moreover, since the training data is constructed from web-crawled sources, there exist inherent low-quality pairs where the face identity and the audio speaker do not match~(e.g., gender mismatch).  Fixed fusion strategies cannot adapt to such per-sample quality variation.  To address this, we propose AMR, a modality fusion module that dynamically assesses per-sample input quality.  Both encoders remain frozen during AMR training; only the modality adapters and the router are trainable.

\subsubsection{AMR Architecture}

Given the audio embedding $\mathbf{e}_a \in \mathbb{R}^{256}$ and the face embedding $\mathbf{e}_f \in \mathbb{R}^{512}$, AMR operates in four steps.

First, two modality adapters process the encoder embeddings into a common 256-dimensional representation space:
\begin{align}
  \mathbf{e}'_a &= \text{ReLU}(\mathbf{W}_{a2} \cdot \text{ReLU}(\mathbf{W}_{a1} \mathbf{e}_a)) \label{eq:audio_adapter} \\
  \mathbf{e}'_f &= \text{ReLU}(\mathbf{W}_{f2} \cdot \text{ReLU}(\mathbf{W}_{f1} \mathbf{e}_f)) \label{eq:face_adapter}
\end{align}
where $\mathbf{W}_{a1} \in \mathbb{R}^{256 \times 256}$, $\mathbf{W}_{a2} \in \mathbb{R}^{256 \times 256}$ for the audio adapter, and $\mathbf{W}_{f1} \in \mathbb{R}^{512 \times 256}$, $\mathbf{W}_{f2} \in \mathbb{R}^{256 \times 256}$ for the face adapter.
The adapted embeddings $\mathbf{e}'_a$ and $\mathbf{e}'_f$ serve exclusively as input to the router for estimating dynamic modality weights; the classification logits are computed directly from the original encoder embeddings via the frozen projection heads.

Second, a trainable router takes the concatenation of the adapted embeddings and estimates dynamic modality weights:
\begin{equation}
  [w_a, w_f] = \text{Softmax}\bigl(\mathbf{W}_{r2} \cdot \text{ReLU}(\mathbf{W}_{r1} [\mathbf{e}'_a; \mathbf{e}'_f])\bigr)
  \label{eq:router}
\end{equation}
where $\mathbf{W}_{r1} \in \mathbb{R}^{512 \times 256}$, $\mathbf{W}_{r2} \in \mathbb{R}^{256 \times 2}$.  A dropout layer with rate 0.2 is applied to the router's hidden layer during training.

Third, the audio logits $\mathbf{l}_a$ and face logits $\mathbf{l}_f$ are obtained from the frozen encoders via their respective projection heads~(Eq.~\ref{eq:audio_logits} and Eq.~\ref{eq:face_logits}).  All encoder parameters and projection weights remain frozen during AMR training.

Finally, the router weights aggregate the modality-specific logits:
\begin{equation}
  \mathbf{l}_{\text{final}} = w_a \cdot \mathbf{l}_a + w_f \cdot \mathbf{l}_f
  \label{eq:fusion}
\end{equation}
The prediction is $\hat{y} = \arg\max(\mathbf{l}_{\text{final}})$.
During inference under audio-only conditions~(protocols P4 and P6), the face pathway is deactivated, and the prediction reduces to $\hat{y} = \arg\max(\mathbf{l}_a)$.

\subsubsection{Modality-Aware Training}
\label{sec:modality_aware}

We design a modality-aware training strategy with four sample types to teach the AMR router how to handle different input conditions.  Since input quality varies across samples, the router must learn to dynamically adjust modality weights based on per-sample reliability.  Moreover, since the task is a \textbf{closed-set identification problem} where all test identities belong to the known 70-speaker set, the router can also learn to detect whether the audio or face input matches a speaker within the closed set, and correspondingly down-weight the inconsistent modality.  Table~\ref{tab:sample_types} summarizes the four training sample types.

\begin{table}[t]
  \caption{Four sample types for AMR training.  Each type simulates a different input quality condition and provides explicit supervision to the router through a target weight distribution.}
  \label{tab:sample_types}
  \footnotesize
  \begin{tabular}{llcc}
    \toprule
    Type & Condition & Prob. & Router Target  \\
    \midrule
    ORIGINAL & Original face + original audio & 0.2 & [0.4, 0.6] \\
    AUDIO\_REPLACE & Original face + external audio & 0.6 & [0.0, 1.0] \\
    FACE\_REPLACE & External face + original audio & 0.1 & [1.0, 0.0] \\
    NO\_FACE & Black face image + original audio & 0.1 & [1.0, 0.0] \\
    \bottomrule
  \end{tabular}
\end{table}

The ORIGINAL type uses matched audio-face pairs with slightly face-biased router targets ($w_f = 0.6$), reflecting the higher accuracy of the face modality in the baseline system.  The AUDIO\_REPLACE type replaces the original audio with an external audio sample from a multi-speaker or noisy source, simulating corrupted audio input.  The router target directs the model to trust the face modality entirely ($w_f = 1.0$).  The FACE\_REPLACE type replaces the original face with an external face from a different identity, simulating a face mismatch.  The router target directs the model to trust the audio modality ($w_a = 1.0$).  The NO\_FACE type replaces the face with a black image, simulating complete face absence; the router target is identical to FACE\_REPLACE.

External audio samples for AUDIO\_REPLACE are drawn from two sources: multi-speaker audio segments identified by speaker diarization, and non-target speaker audio from VoxCeleb.  External face images for FACE\_REPLACE are drawn from the CelebA dataset~\cite{celeba}.

\subsubsection{Loss Function}

The total loss combines cross-entropy classification loss with KL divergence on the router weights:
\begin{align}
  \mathcal{L}_{\text{total}} &= \mathcal{L}_{\text{CE}} + \lambda \cdot \mathcal{L}_{\text{KL}} \label{eq:total_loss} \\
  \mathcal{L}_{\text{CE}} &= \text{CrossEntropy}(\mathbf{l}_{\text{final}},\, y) \label{eq:ce_loss} \\
  \mathcal{L}_{\text{KL}} &= \text{KL}\bigl(\mathbf{w}_{\text{target}} \,\|\, \mathbf{w}\bigr) \label{eq:kl_loss}
\end{align}
where $y$ is the ground-truth label, $\mathbf{w} = [w_a, w_f]$ are the router output weights, $\mathbf{w}_{\text{target}}$ is the target distribution from Table~\ref{tab:sample_types}, and $\lambda = 1.0$ balances the two terms.  The cross-entropy loss drives the fused prediction toward the correct identity.  The KL divergence loss provides direct supervision to the AMR router, teaching the router to assign appropriate modality weights based on input quality.

\subsubsection{Training Configuration}

AMR training uses the Adam optimizer with learning rate $2 \times 10^{-3}$ and weight decay $10^{-4}$. The batch size is 256 per GPU. Training runs for 30 epochs.  The training data consists of audio-face pairs from the MAV-Celeb dataset~\cite{mavceleb}, expanded by pairing each audio sample with multiple face images from the same speaker~(up to 50 face crops per speaker), yielding approximately 499K training pairs.

\subsection{Data Preparation Pipeline}
\label{sec:data}

\subsubsection{Audio Pipeline}
\label{sec:audio_data}
\begin{figure}[t]
  \centering
  \includegraphics[width=0.95\columnwidth]{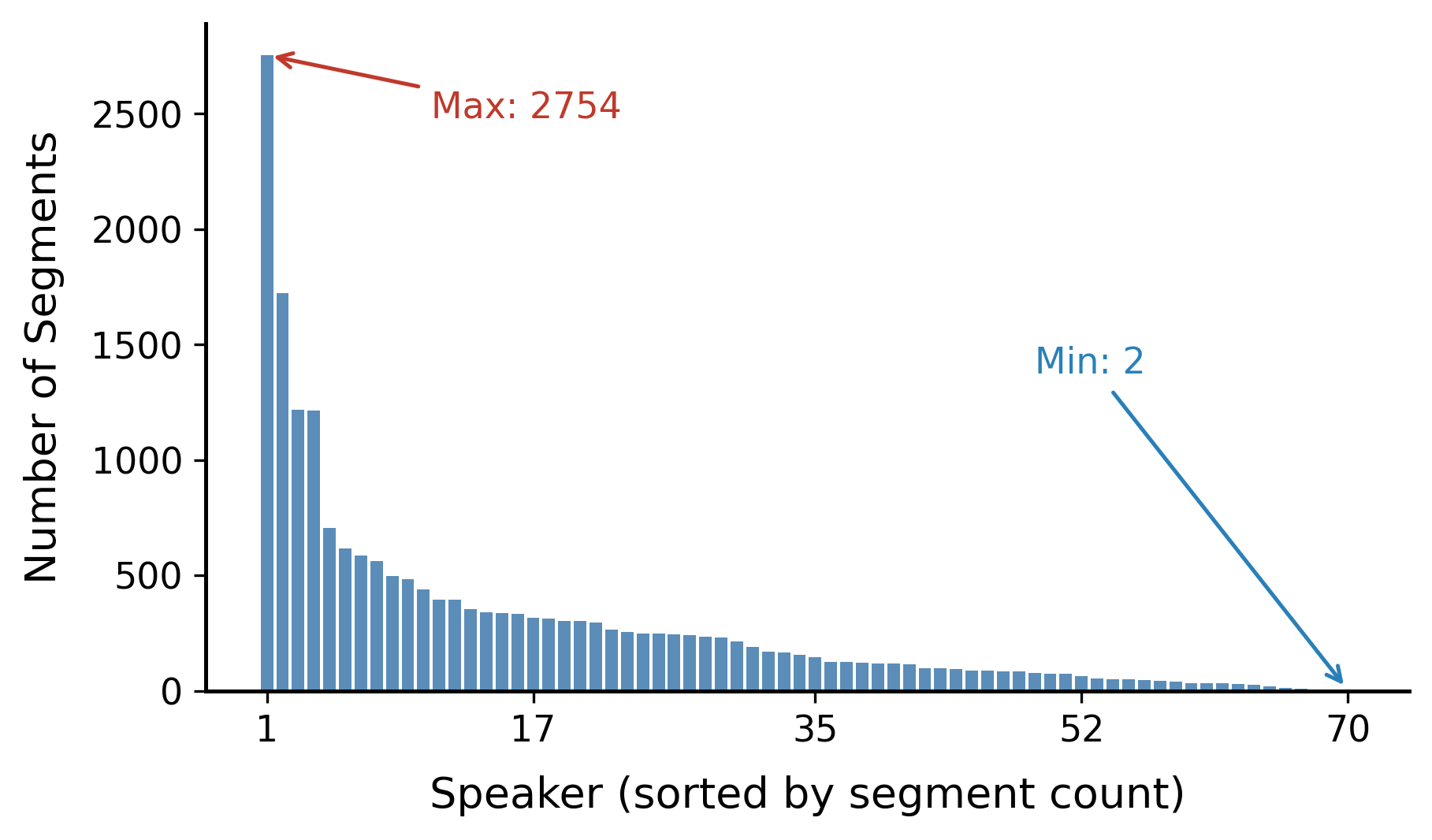}
  \caption{Per-speaker sample distribution after speech segmentation and multi-speaker filtering. }
  \label{fig:data_distribution}
  \Description{Bar chart showing the number of audio segments per speaker, revealing a long-tailed distribution with significant class imbalance.}
\end{figure}

\begin{sloppypar}
The training audio is sourced from web-crawled MAV-Celeb videos, which present three quality challenges:~(1)~background noise and overlapping speech corrupt the acoustic signal;~(2)~multi-speaker segments contain mixed identities;~(3)~extreme per-speaker sample imbalance creates a long-tailed distribution~(Figure~\ref{fig:data_distribution}).
\end{sloppypar}

To address these issues, we apply a multi-stage pipeline comprising speech segmentation, multi-speaker filtering, speaker similarity cleaning, class balancing, and TTS augmentation.

\textbf{Speech segmentation and filtering.}
Raw audio is segmented using FireRedVAD~\cite{fireredvad} voice activity detection, retaining speech regions longer than 1.0~seconds.  Pyannote speaker diarization~\cite{pyannote} then identifies and removes segments containing multiple speakers.
\begin{sloppypar}
\textbf{Quality filtering and manual annotation.}
A pretrained ResNet293 speaker model computes embeddings for all segments.  Segments whose cosine similarity to the target speaker's average embedding falls below a given threshold are removed as likely mislabels.  The remaining segments undergo manual verification~(up to 200 per speaker) to confirm speaker identity.
\end{sloppypar}

\textbf{Class balancing and augmentation.}
Per-speaker sample counts are balanced via under-sampling majority classes and over-sampling minority classes.  Additional training diversity is introduced through TTS synthesis~(CosyVoice3~\cite{cosyvoice3} and VoxCPM2~\cite{voxcpm}; see Section~\ref{sec:audio_training}) and background noise augmentation~(MUSAN dataset~\cite{musan}, 50\% probability, SNR sampled from $[0, 20]$~dB during training).

\subsubsection{Face Pipeline}
\label{sec:face_data}

Face training data comes from three sources: web-crawled YouTube videos~(detected and aligned using a face detector), challenge-provided MAV-Celeb crops~(already aligned), and Nano Banana synthetic data.  Manual inspection and quality filtering were applied across all sources to remove misaligned, low-quality, and identity-inconsistent samples.  All face images are resized to $112 \times 112$ and normalized with a mean of 0.5 and a standard deviation of 0.5.

\section{Experiments}
\label{sec:experiments}

\subsection{Dataset and Evaluation Protocol}
\label{sec:dataset}

The POLY-SIM 2026 dataset is based on the MAV-Celeb dataset~\cite{mavceleb}, which involves 70 bilingual speakers.  Table~\ref{tab:dataset} summarizes dataset statistics.

\begin{table}[t]
  \caption{MAV-Celeb dataset statistics for the English--Urdu language pair used in POLY-SIM 2026.}
  \label{tab:dataset}
  \begin{tabular}{lccc}
    \toprule
    Split & English & Urdu & Total \\
    \midrule
    Training samples & 4,039 & 9,304 & 13,343 \\
    Evaluation samples & 1,521 & 1,623 & 3,144 \\
    \bottomrule
  \end{tabular}
\end{table}

The four evaluation protocols test different conditions:
\begin{itemize}
  \item \textbf{P3}~(In-language multimodal): Both audio and face are available at test time.  Training and testing use English.
  \item \textbf{P4}~(In-language audio-only): Only audio is available at test time.  Training and testing use English.
  \item \textbf{P5}~(Cross-lingual multimodal): Both audio and face are available.  Training uses English; testing uses Urdu.
  \item \textbf{P6}~(Cross-lingual audio-only): Only audio is available.  Training uses English; testing uses Urdu.
\end{itemize}

The evaluation metric is classification accuracy~(P-accuracy): the proportion of test samples for which the predicted identity matches the ground truth.  The overall score is the average accuracy across all four protocols.

\subsection{Main Results}
\begin{table}[t]
  \caption{Comparison with the FOP baseline on the POLY-SIM 2026 evaluation set.  Accuracy~(\%) is reported for each protocol.}
  \label{tab:results}
  \begin{tabular}{lccccc}
    \toprule
    System & P3 & P4 & P5 & P6 & Avg. \\
    \midrule
    Baseline~(FOP) & 97.44 & 37.75 & 98.48 & 31.7 & 66.34 \\
    \textbf{Ours~(AMR)} & \textbf{99.93} & \textbf{97.50} & \textbf{100.00} & \textbf{98.83} & \textbf{99.07} \\
    \bottomrule
  \end{tabular}
\end{table}

Table~\ref{tab:results} compares our system against the FOP baseline~\cite{fop} provided by the challenge organizers.
Our system achieves 99.93\% on P3~(only 1 error out of 1,521 samples), 100.00\% on P5~(perfect accuracy on all 1,623 Urdu samples), and audio-only performance on P4~(97.50\%) and P6~(98.83\%).  The average accuracy of 99.07\% represents a 32.73\% improvement over the FOP baseline.

\subsection{Analysis}

\textbf{Overall performance.}
The improvement over the FOP baseline is most pronounced on audio-only protocols: P4 improves from 37.75\% to 97.50\%~(+59.75) and P6 from 31.70\% to 98.83\%~(+67.13).  On multimodal protocols, P3 improves from 97.44\% to 99.93\% and P5 from 98.48\% to 100.00\%.  The low audio-only baseline scores indicate that the FOP architecture is heavily dependent on the face modality and cannot gracefully handle missing faces.  Our independently optimized audio encoder achieves 97.50\% on P4 and 98.83\% on P6 as a standalone classifier~(Table~\ref{tab:amr_ablation}).  Notably, the audio encoder is fine-tuned on English data only, yet achieves 98.83\% on Urdu audio~(P6), demonstrating strong cross-lingual transferability of the W2V-BERT~2.0 pretrained encoder.

\textbf{Fusion benefit.}
The face-only model achieves 99.93\% on P3 and 99.63\% on P5~(Table~\ref{tab:amr_ablation}), indicating that the face modality alone is highly discriminative for this 70-speaker task.  However, we observe that the evaluation set contains a non-trivial number of low-quality audio-face pairs, including cases where the face identity and the audio speaker \textbf{do not match in gender}~(e.g., id00320, id00719, id00125, id00273 in the English evaluation set, and id00858, id01353 in the Urdu evaluation set).  Such noisy samples make it unreliable to rely on a single modality.  On P5, AMR recovers all 6 face-only errors by leveraging the audio pathway, achieving 100.00\%.  On P3, face-only and full AMR both achieve 99.93\%, with only 1 error where both modalities predict incorrectly.

\subsection{Ablation Studies}

To understand the contribution of individual components, we conduct ablation studies on both the AMR fusion module and the audio encoder training stages.

\subsubsection{AMR Component Ablation}

Table~\ref{tab:amr_ablation} shows the ablation results.  We start from the full AMR training as described in Section~\ref{sec:amr} and progressively remove key components.  All removals affect only the multimodal protocols~(P3, P5), while the audio-only protocols~(P4, P6) remain unchanged.  Removing the KL divergence loss~(i.e., setting $\lambda = 0$ in Eq.~\ref{eq:total_loss}) drops P5 from 100.00\% to 99.75\% and P3 from 99.93\% to 99.80\%, confirming that explicit router supervision is critical for learning appropriate modality weights.  Further removing modality-aware training~(i.e., training with only ORIGINAL samples using $\mathcal{L}_{\text{CE}}$ in Eq.~\ref{eq:ce_loss}) degrades P5 from 99.75\% to 99.38\% and P3 from 99.80\% to 99.74\%, showing that the four-sample-type strategy is essential for handling diverse input conditions.

\begin{table}[t]
  \caption{Ablation study on AMR components.  Accuracy~(\%) reported on the POLY-SIM evaluation set.}
  \label{tab:amr_ablation}
  \footnotesize
  \begin{tabular}{lccccc}
    \toprule
    Configuration & P3 & P4 & P5 & P6 & Avg \\
    \midrule
    Audio only & - & 97.50 & - & 98.83 & - \\
    Face only & {99.93} & - & {99.63} & - & - \\
    \midrule
    \textbf{AMR~(Full)} & \textbf{99.93} & \textbf{97.50} & \textbf{100.00} & \textbf{98.83} & \textbf{99.07} \\
    \quad $-$ KL loss~($\lambda\!=\!0$) & 99.80 & 97.50 & 99.75 & 98.83 & 98.97 \\
    \quad $-$ KL \& modality-aware & 99.74 & 97.50 & 99.38 & 98.83 & 98.86 \\
    \bottomrule
  \end{tabular}
\end{table}

\subsubsection{Audio Training Stage Ablation}

Table~\ref{tab:audio_ablation} shows the contribution of each stage in the three-stage progressive audio encoder training. The ablation confirms that each training stage contributes to improved audio-only accuracy.  Stage~1 establishes a baseline of 93.00\% by training the projection head while keeping the SSL frontend and MFA frozen.  Stage~2 unfreezes the MFA layers, enabling multi-scale feature aggregation to adapt to the target speakers and improving the average to 95.90\%.  Stage~3 adds TTS augmentation, exposing the model to synthesized speech with diverse prosodic characteristics, yielding a further improvement to 98.17\%.

\begin{table}[t]
  \caption{Ablation study on audio training stages.  Audio-only accuracy~(\%) on validation set reported.}
  \label{tab:audio_ablation}
  \begin{tabular}{lccc}
    \toprule
    Stages & P4 & P6 & Avg \\
    \midrule
    Stage 1 only & 92.02 & 93.99 & 93.00 \\
    Stage 1+2 & 95.73 & 96.06 & 95.90 \\
    Stage 1+2+3~(Full) & \textbf{97.50} & \textbf{98.83} & \textbf{98.17} \\
    \bottomrule
  \end{tabular}
\end{table}

\section{Conclusion}
\label{sec:conclusion}

This paper addresses the challenges of missing modality and unreliable input quality in multimodal polyglot speaker identification.  We propose a system built upon AMR, a modality fusion module that dynamically assesses per-sample input quality.  AMR employs two modality adapters to process the embeddings from an audio encoder~(W2V-BERT 2.0 with MFA), a face encoder~(IResNet-18), and a trainable router estimates dynamic modality weights to aggregate the modality-specific logits for the final prediction.  The AMR module is trained via a modality-aware strategy with KL divergence supervision while both encoders remain frozen.

Experiments on the POLY-SIM 2026 evaluation set demonstrate the effectiveness of the proposed system.  The system achieves 99.07\% average accuracy across four protocols, surpassing the FOP baseline by 32.73\%.  

A current limitation is that the system has only been evaluated on a single bilingual pair~(English--Urdu) with 70 speakers.  Extending to more languages, larger speaker populations and additional modalities could be a next step.  The AMR paradigm demonstrates that dynamically routed fusion is a practical approach for real-world multimodal speaker identification systems under the closed-set setting.

\begin{acks}
This work was conducted as part of the POLY-SIM 2026 Grand Challenge at ACM Multimedia 2026. The authors thank the organizers for providing the MAV-Celeb dataset and evaluation infrastructure, and Weishan Li for assistance with dataset cleaning.
\end{acks}

\section*{Ethics and Privacy Statement}

This work uses the MAV-Celeb dataset, which contains audio-visual data of public figures collected from YouTube. While the source videos are publicly available, the use of biometric data raises privacy considerations. No additional personal data beyond publicly accessible content was used. The proposed speaker recognition models should be deployed only with appropriate user consent and privacy safeguards. The authors advocate for responsible use of biometric identification technologies.

\bibliographystyle{ACM-Reference-Format}
\bibliography{references}

\end{document}